\documentclass{article}

\usepackage[preprint]{corl_2024} % Uncomment for pre-prints (e.g., arxiv); This is like ``final'', but will remove the CORL footnote.
\usepackage[numbers]{natbib}
\usepackage{multicol}
\usepackage{amsmath, amssymb}
\usepackage{graphicx}
\usepackage{comment}
\usepackage{booktabs}
\usepackage[font=small,labelfont=bf]{caption}
\usepackage{soul}
\usepackage{tabularx}
\usepackage{wrapfig}
\usepackage{bm}
\usepackage{cleveref} 

\usepackage{xcolor}
\definecolor{ourblue}{rgb}{0.368,0.507,0.71}
\definecolor{ourorange}{rgb}{0.881,0.611,0.142}
\definecolor{ourgreen}{rgb}{0.56,0.692,0.195}
\definecolor{ourred}{rgb}{0.923,0.386,0.209}
\definecolor{ourviolet}{rgb}{0.528,0.471,0.701}
\definecolor{ourbrown}{rgb}{0.772,0.432,0.102}
\definecolor{ourlightblue}{rgb}{0.364,0.619,0.782}
\definecolor{ourdarkgreen}{rgb}{0.572,0.586,0.}

\definecolor{ourcyan2}{rgb}{0.125,0.722,0.804}
\definecolor{ourred2}{rgb}{0.863,0.184,0.047}
\definecolor{ouryellow2}{cmyk}{0,0.16,1.0,0.07}
\definecolor{ourviolet2}{cmyk}{0.55,0.56,0,0.47}
\definecolor{ourorange2}{cmyk}{0,0.46,0.89,0.11}

\hypersetup{
	colorlinks=true,       % false: boxed links; true: colored links
	linkcolor=ourviolet,        % color of internal links
	citecolor=ourblue,        % color of links to bibliography
	filecolor=ourblue,     % color of file links
	urlcolor=ourviolet
}

\newcommand{\R}{\mathbb{R}}
\newcommand{\eg}{\emph{e.\,g.},{ }}

\newcommand{\prb}{\mathrm{prb}}

\newcommand{\Figref}[1]{Figure~\ref{#1}}  % beginning of sentence
\newcommand{\figref}[1]{Figure~\ref{#1}}    % somewhere

\def\eqref#1{(\ref{#1})}

\title{Learning deformable linear object dynamics\\ from a single trajectory}

\author{
  Shamil Mamedov\\
  The MECO Research Team \\
  KU Leuven, 3000 Leuven, Belgium \\
  \texttt{shamil.mamedov@kuleuven.be} \\
  \And
  A.~René Geist \\
  \texttt{rene.geist@dsme.rwth-aachen.de}
  \And
  Ruan  Viljoen\\
  The MECO Research Team \\
  KU Leuven, 3000 Leuven, Belgium \\
  \texttt{ruan.viljoen@kuleuven.be} \\
  \And
  Sebastian Trimpe\\
  The Institute for DSME\\
   RWTH Aachen University, 52068 Aachen, Germany\\
   \texttt{ trimpe@dsme.rwth-aachen.de} \\
   \And
  Jan Swevers \\
  The MECO Research Team \\
  KU Leuven, 3000 Leuven, Belgium \\
  \texttt{jan.swevers@kuleuven.be} \\
}

\begin{document}
\maketitle

%===============================================================================

\begin{abstract}
\begin{comment}
        The dynamic manipulation of deformable objects poses a significant challenge in robotics. 
    While model-based approaches for controlling such objects hold significant potential, their effectiveness hinges on the availability of an accurate and computationally efficient dynamics model.
    This paper focuses on sample-efficient learning of  models for capturing the dynamic behavior of deformable linear objects (DLOs).
    Inspired by pseudo-rigid body and rigid FEM methods, we present a physics-informed neural ODE that approximates a DLO as a serial chain of rigid bodies interconnected by passive elastic joints.
    However, unlike the traditional uniform discretization and linear spring-damper joints, our approach involves learning-based discretization and  nonlinear elastic joints that characterize interaction forces via a neural network.
    Through experiments involving two DLOs with markedly different physical properties, we demonstrate the efficacy of the proposed model in accurately predicting DLO motion.
    \newline The project code and data are available at: \textit{at this stage attached as a supplementary material.}
\end{comment}
    The manipulation of deformable linear objects (DLOs) via model-based control requires an accurate and computationally efficient dynamics model. Yet, data-driven DLO dynamics models require large training data sets while their predictions often do not generalize, whereas physics-based models rely on good approximations of physical phenomena and often lack accuracy. To address these challenges, we propose a physics-informed neural ODE capable of predicting agile movements with significantly less data and hyper-parameter tuning. In particular, we model DLOs as serial chains of rigid bodies interconnected by passive elastic joints in which interaction forces are predicted by neural networks. The proposed model accurately predicts the motion of an robotically-actuated aluminium rod and an elastic foam cylinder after being trained on only thirty seconds of data.
    
    The project code and data are available at: \url{https://tinyurl.com/neuralprba}
\end{abstract}

\keywords{Deformable linear objects, Physics-informed machine learning} 

%===============================================================================

\section{Introduction}

\begin{comment}
\begin{wrapfigure}{r}{0.5\textwidth}
    \vspace{-6mm}
  \begin{center}
    \captionsetup{font=small}
    \includegraphics[width=0.47\textwidth]{images/overview_v5.pdf}
    \caption{Overview on the proposed physics-informed neural ODE for modeling DLO dynamics. We discretize a DLO as a chain of pseudo-rigid bodies whose hybrid dynamics are derived via the arcticulated body algorithm. The interaction forces acting between the pseudo-bodies are learned by neural networks.}
    \label{fig:DlOs}
  \end{center}
  \vspace{-5mm}
\end{wrapfigure}
\end{comment}

\begin{figure}[t]
  \begin{center}
    \captionsetup{font=small}
    \includegraphics[width=1.\textwidth]{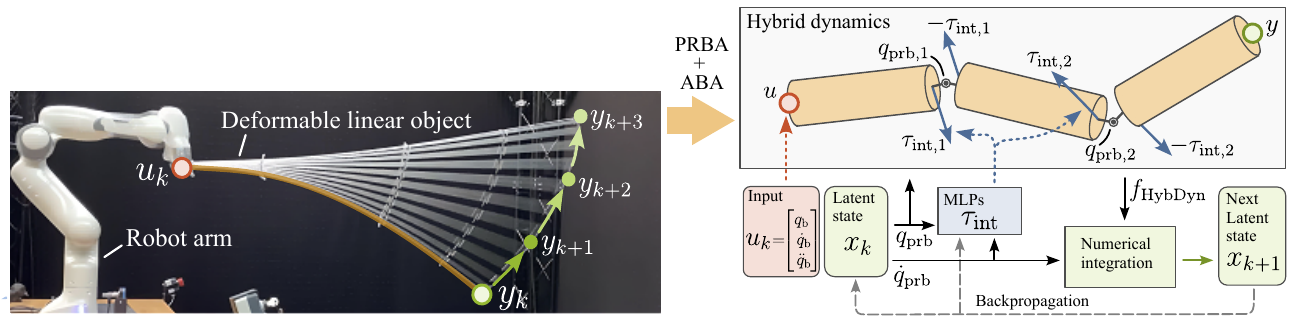}
    \caption{Overview on the proposed physics-informed neural ODE for modeling DLO dynamics. We discretize a DLO as a chain of pseudo-rigid bodies whose hybrid dynamics are derived via the articulated body algorithm. The interaction forces acting between the pseudo-bodies are learned by neural networks.}
    \label{fig:DlOs}
  \end{center}
  \vspace{-5mm}
\end{figure}

Manipulating deformable objects poses a significant challenge within the field of robotics. 
Model-based control approaches have proven effective for numerous tasks. However, their effectiveness relies on computationally efficient and accurate dynamics models.
%Acquiring such a model is challenging, as deformable object dynamics are infinite-dimensional and form partial differential equations \cite{arriola2020DOmodelingSurvey, sanchez2018robotic}.
Acquiring such dynamics models is challenging as the interactions inside mass continua naturally give rise to partial differential equations \cite{arriola2020DOmodelingSurvey, sanchez2018robotic}.
This work addresses the sample-efficient modeling of deformable linear objects (DLOs), such as cables, ropes and rods, for manipulation. 
In contrast to existing literature that focuses on quasi-static behavior and assumes access to the full state \citep{yan2020biLSTMRope, yang2021biLSTM_IN, yu2022RBF}, our interest lies in highly dynamic DLO motions where only partial observations are available. More specifically, we aim to predict the motion of one end of a DLO over time, given estimates of positions, velocities, and accelerations of the other actuated end, as shown in \figref{fig:DlOs}. Such models are relevant for various applications, including the handling of flexible objects in manufacturing and construction. 

A fundamental challenge when modeling DLOs lies in selecting %an appropriate and computationally tractable state representation. 
a suitable state representation.
Drawing inspiration from pseudo-rigid body (PRB) approaches \cite{bauchau2011flexible} and rigid FEM \cite{wittbrodt2007RFEM}, we model DLOs as chains of rigid bodies connected via passive elastic joints \cite{mamedov2023FEIN}. 
Consequently, the state variables are the positions and velocities of these passive joints. 
In contrast to \cite{mamedov2023FEIN} where the dynamics are modeled via black-box recurrent networks, %we propose a novel approach that combines a physics model with neural networks, aiming to obtain a sample-efficient yet expressive continuous-time dynamics model.
we combine a continuous-time physics model with neural networks to obtain a sample-efficient yet expressive model.
The advantages of our model are:
\begin{enumerate}
    %\item \emph{Sample-efficiency}: can be trained with just a single trajectory containing thirty seconds of data;
    \item \emph{Sample-efficiency}: yields accurate predictions after training on thirty seconds of data;
    \item \emph{Computational efficiency}: yields  a low-dimensional, physically meaningful state representation, facilitating faster numerical integration compared to analytical physics models; 
    %\item \emph{Versatility}: capable of learning not only nonlinear elasticity but also plastic material deformations from data.
    \item \emph{Versatility}: capable of learning nonlinear elastic and plastic material deformations.
\end{enumerate}

% Additionally, we adopt a straightforward method for estimating initial states, which avoids the need for training a recognition model as in other works \cite{mamedov2023FEIN, preiss2022trackingPoolNoodle}. 

%We demonstrate through two robot experiments on DLOs with different material properties that the proposed model significantly outperforms black-box and physics-based models in the small data regime.

\begin{comment}
\begin{wrapfigure}{t}{0.5\textwidth}
    \vspace{-6mm}
  \begin{center}
    \captionsetup{font=small}
    \includegraphics[width=0.48\textwidth]{images/overview_v5.pdf}
    \caption{Overview on the proposed physics-informed neural ODE for modeling DLO dynamics. We discretize a DLO as a chain of pseudo-rigid bodies whose hybrid dynamics are derived via the arcticulated body algorithm. The interaction forces acting between the pseudo-bodies are learned by neural networks.}
    \label{fig:DlOs}
  \end{center}
  \vspace{-5mm}
\end{wrapfigure}
\end{comment}

%===============================================================================
\section{Related work} \label{sec:rel_work}
Various communities, including flexible robotics, soft robotics, robotic manipulation, and computer graphics, have proposed methods for modeling the dynamics of DLOs, ranging from analytical to data-driven approaches. Our discussion of related work focuses on the key aspects of our contribution: learning DLO dynamics from data, defining a suitable state representation, and estimating latent states from input-output data. For a comprehensive review of DLO modeling, interested readers are referred to \cite{arriola2020DOmodelingSurvey, sanchez2018robotic, lv2017physically, armanini2023soft}.

\subsection{Learning DLO dynamics}
The predominant focus in modeling DLOs has been directed towards predicting their slow quasi-static behavior \citep{yan2020biLSTMRope, yang2021biLSTM_IN, yu2022RBF}. For instance, \citet{yu2022RBF} employed radial basis functions to model a DLO at a velocity level (differential kinematics) using data generated in simulation, subsequently fine-tuning the model online during manipulation. In another approach, \citet{yan2020biLSTMRope} utilized a bidirectional long-short term memory network (bi-LSTM) to model a rope at a position level (kinematics), propagating latent states spatially rather than temporally from one end to another. \citet{yang2021biLSTM_IN} combined bi-LSTMs with interaction networks \cite{battaglia2016interaction} to model a cable fixed at one end and manipulated by a robot at the other.

Among the few papers aiming to learn dynamics models of DLOs, \cite{mamedov2023FEIN} and \cite{preiss2022trackingPoolNoodle} stand closest to our approach. \citet{preiss2022trackingPoolNoodle} utilized an LSTM to predict DLO endpoint motion based on pitch and yaw angles of the robot's end-effector as inputs, learning a large-dimensional latent state from approximately 200 trajectories. \citet{mamedov2023FEIN} proposed combining black-box discrete-time dynamics models with a physics-informed encoder and decoder, aiming for an interpretable latent state. While both these approaches yield good results, their major limitation is the need for a significant amount of data to learn good parameter values.
In contrast, our approach substantially improves both sample efficiency and generalizibility by employing a continuous-time hybrid model that integrates more physics-based knowledge.  
In the results section, we compare our method against that of \cite{mamedov2023FEIN} which performs similarly to \cite{preiss2022trackingPoolNoodle}.

\subsection{State representation for DLOs}
Two common approaches for representing states of DLOs are particle-based and spatial discretization methods. 
The particle-based representation, where a DLO is approximated by a collection of particles with latent states being position and velocity of each particle, stands out as a simpler approach for learning dynamics from data. This is notably due to the ease of measuring positions and velocities of points along a DLO using cameras or dedicated sensors \cite{yan2020biLSTMRope, yu2022RBF}. Additionally, certain methods like graph neural networks inherently adopt a particle-based approach \cite{li2019propNet}. However, this representation has a limitation: over prolonged rollouts, particles may diverge, leading to predictions that lack physical consistency as illustrated in \cite{yang2021biLSTM_IN}.

In contrast, spatial discretization methods represent DLOs as a series of straight segments (aka bodies or links) that are
connected by joints \cite{yang2021biLSTM_IN, moberg2014EFJ, bergou2008discreterRods}. Such a representation is subject to kinematic constraints which ensure that predictions remain physically consistent. 
%Within this approach, common rotation parameterizations include Euler angles, quaternions, and position-twist angle pairs \cite{lefevre20174}.
%
In this paper, we leverage the spatial discretization method and approximate a DLO as a chain of rigid bodies, employing relative angles and angular velocities as the its state \cite{wittbrodt2007RFEM, moberg2014EFJ}. An important aspect of our method is the optimization of spatial discretization (rigid body lengths) in a data-driven manner, leading to a compact and efficient state representation.

\subsection{Estimating latent states of DLO from input-output data}
Latent dynamic models require an initial state and a sequence of inputs to predict the motion of DLOs. However, in contrast to \cite{yang2021biLSTM_IN}, direct state measurements are often not available and must be estimated from input-output data. 
This poses a challenge, as classic state estimators, such as the Kalman filter and its variations, requires a dynamic model to estimate the states. This results in a chicken-and-egg problem: state estimates are needed to learn the dynamics model, but the dynamic model is required to find the state estimates.
In the machine learning literature, methods addressing this challenge fall under the category of \textit{recognition models}. These models infer the initial state by leveraging a history of inputs and observations backward in time \cite{buisson2022recognitionModels, chen2018NODE} or forward in time \cite{ beintema2022DeepSubspace}. 

In the context of DLO modeling, \cite{yan2020biLSTMRope, yang2021biLSTM_IN, yu2022RBF} estimated states directly from camera images, \cite{preiss2022trackingPoolNoodle} initialized the latent state of an LSTM with zeros, and \cite{mamedov2023FEIN} trained a physics-informed neural network recognition model to obtain latent states.
For small datasets, recognition models are not useful, as they require substantial amounts of data and are prone to overfitting. Therefore, we adopt a more lightweight approach that circumvents the need to train a recognition model by jointly optimizing the initial states and model parameters.

%===============================================================================
\section{Problem Formulation} \label{sec:prob_statement}

\begin{wrapfigure}{r}{0.45\textwidth}
    \vspace{-9mm}
  \begin{center}
    \captionsetup{font=small}
    \includegraphics[width=0.45\textwidth]{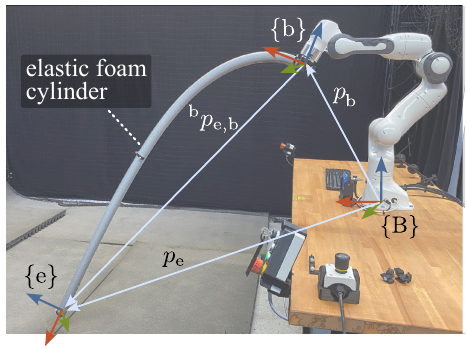} 
    \caption{The notation used throughout the paper. $\{\mathrm{B}\}$ is the inertial frame attached to the base of Panda, $\{\mathrm{b}\}$ and $\{\mathrm{e}\}$ are frames attached to the start and end of a DLO, respectively.} \label{fig:notation}
  \end{center}
  \vspace{-4mm}
\end{wrapfigure}
In this section, we formally define the problem addressed in this paper, beginning with an outline of the experimental setup and the main notation used, as depicted in \figref{fig:notation}. 
The inertial coordinate frame is represented by $\{B\}$, while the frames fixed to the start and end of the DLO are denoted as $\{\mathrm{b}\}$ and $\{\mathrm{e}\}$, respectively.  Our dataset, $\mathcal{D}$, consists of a series of outputs $y=[p_\mathrm{e}^\top\ \dot p_\mathrm{e}^\top]^\top \in \R^{6}$
% \footnote{$[a, b] \doteq [a^\top b^\top]^\top$ denotes the concatenation of two column vectors.}
, which include the position $p_\mathrm{e}$ and linear velocity $\dot p_\mathrm{e}$ of $\{\mathrm{e}\}$ relative to $\{B\}$, and corresponding inputs $u=[q_\mathrm{b}^\top\ \dot q_\mathrm{b}^\top\ \ddot q_\mathrm{b}^\top]^\top \in \R^{18}$, which include the pose $q_\mathrm{b}$, velocity $\dot q_\mathrm{b}$, and acceleration $\ddot q_\mathrm{b}$ of $\{\mathrm{b}\}$ relative to $\{B\}$. The latent state of the DLO is denoted by 
% $x=[q_\mathrm{prb}, \dot q_{\mathrm{prb}}] \in \R^{n_x}$
$x \in \R^{n_x}$. Building on this notation, we define the the problem addressed in this work as follows:

\emph{
Given a small dataset $\mathcal{D}_\mathrm{train}$, learn a latent dynamical model  $\dot x = f(x, u)$ together with a decoder $y=h(x,u)$ intended to predict the future end-point position and velocities $y_{k+1}, y_{k+2}, \dots, y_{k+N}$, 
given: a) the current end-point position and velocity $y_k$, and b) a sequence of future inputs $u_k, u_{k+1},\dots, u_{k+N}$. 
}

The term ``small dataset'' refers specifically to a single trajectory of a reasonable duration, typically less than a minute. 

% \begin{figure}[t]
%     \captionsetup{font=small}
%     \centering
%     %\includegraphics[width=\linewidth, trim={8.6cm 0 0 0}, clip]{images/notation.pdf}
%     \includegraphics[width=0.48\linewidth]{images/overview_v1.pdf}
%     \caption{The notation used throughout the paper. $\{\mathrm{B}\}$ is the inertial frame attached to the base of Panda, $\{\mathrm{b}\}$ and $\{\mathrm{e}\}$ are frames attached to the start and end of a DLO, respectively.}
%     \label{fig:notation}
%     \vspace{-5mm}
% \end{figure}

%===============================================================================
\section{Method} \label{sec:method}
DLOs are infinite-dimensional systems, with their true dynamics governed by partial differential equations. The complexity of learning these equations, combined with their slow integration time, renders them impractical for real-time control.  To develop a computationally efficient approximate dynamical model $\dot x = f(x, u)$ for DLOs, we approximate spatially continuous DLOs as a serial chain of rigid bodies \cite{mamedov2023FEIN, moberg2014EFJ}, an approach commonly known as pseudo-rigid body approximation (PRBA). 
By reformulating DLO dynamics as rigid body dynamics, we can leverage efficient existing algorithms such as the articulated body algorithm \cite{featherstone2014RBAlgos}.  Additionally, we obtain a decoder $h(x,u)$ based on forward kinematics, which maps the states $x$ and input $u$ to the corresponding output $y$.

Classical PRBA, which typically employs fixed discretization (either uniform or based on heuristics) and linear interaction torques, can accurately approximate DLO dynamics if fine discretizations are used \cite{wittbrodt2007RFEM}.  
However, such discretizations leads to high-dimensional continuous-time models whose slow numerical integration is unsuited for real-time control. 
We extend classical PRBA with the goal of achieving similar accuracy with a much coarser discretization, leading to a more computationally efficient model. This we achieve by optimizing the discretization as well as the interaction forces by a neural network.
The developed approach, termed Neural PRBA (NPRBA), is illustrated in \figref{fig:DlOs}. 
The following subsections provide a detailed description of NPRBA.
% \ref{subsec: kin} introduces the kinematics of NPRBA and the resulting decoder equation, \ref{subsec: dyn} derives the dynamics of NPRBA, \ref{subsec: state_est} introduces an approach for obtaining latent states during training and testing, and \ref{subsec: model_train} discusses the loss function used for optimizing parameters from data.

\subsection{Kinematics of NPRBA} \label{subsec: kin}
\Figref{fig:DlOs} shows a typical PRBA of a DLO. The pose of the $i$-th body in the chain is described via a body-fixed coordinate frame $\{i\}$. The transformation between consecutive bodies depends on the pseudo-link length and a passive joint.  While a general joint may possess six degrees of freedom (DOF) —three translational and three rotational—considering inextensible deformable objects, only rotational degrees of freedom are sufficient. Additionally, we disregard twisting owing to the significantly higher twisting stiffness compared to bending in the DLO. Thereby, we represent the transformation between bodies using a 2 DOF rotational elastic joint parameterized with Euler angles, denoted by $q_{\prb,i}$.  

The pose of the DLO's start, i.e. the $\{\mathrm{b}\}$ frame, is described using a 6 DOF joint with Euler angle parameterization for rotation: $q_\mathrm{b} = [p_\mathrm{b}^\top\ \Psi_\mathrm{b}^\top]^\top \in \R^{6}$.
Position and velocity of the DLO's end can be computed via forward and differential forward kinematics of the serial chain \cite{featherstone2014RBAlgos} as 
\begin{align} \label{eq:fk}
    p_\mathrm{e} &= \mathrm{fk}(q_\mathrm{b}, q_\prb, \theta_{\mathrm{kin}}),\ \dot p_{\text{e}} = 
    \begin{bmatrix}
        \frac{\partial \mathrm{fk}}{\partial q_{\mathrm{b}}}  & \frac{\partial \mathrm{fk}}{\partial q_{\prb}} 
    \end{bmatrix}
    \begin{bmatrix}
        \dot q_{\mathrm{b}} \\ \dot q_{\prb}
    \end{bmatrix}, \\
    y &\doteq [p_\mathrm{e}^\top\: \dot p_\mathrm{e}^\top]^\top = h(x, u, \theta_{\mathrm{kin}}), \label{eq: fk_decoder}
\end{align}
where $\mathrm{fk}(\cdot)$ is a forward kinematics map,  $\theta_{\mathrm{kin}}$ is a vector of kinematics parameters comprising lengths of each body, $q_\prb$ and $\dot q_\prb$ are the generalized positions and velocities, respectively, of passive elastic joints, and $x \doteq [q_\prb^\top\ \dot q_\prb^\top]^\top$ is a state of a DLO. Given the chosen state representation, the position and velocity of any point along the DLO can be computed using forward kinematics.

\subsection{Dynamics of NPRBA} \label{subsec: dyn}
In the simulation of rigid robots, forward dynamics typically establishes the relationship between torques applied to the robot's joints and the resulting joint accelerations \cite{featherstone2014RBAlgos}. In our setting, we cannot directly measure the applied torques that manipulate the DLO. Instead, we measure positions, velocities, and accelerations. Hence, we adopt a hybrid dynamics approach \cite{featherstone2014RBAlgos}, which combines forward dynamics for some joints and inverse dynamics for others. Without detailing the hybrid dynamics algorithm, we denote the DLO's second order dynamics as:
\begin{align} \label{eq:nprba_dynamics}
    \ddot q_\prb = f_\mathrm{HybDyn}(q_\prb, \dot q_\prb, \tau_\mathrm{int}, u, \theta),
\end{align}
where $\tau_\mathrm{int}$ defines interaction forces between bodies and $\theta = [\theta_{\mathrm{kin}}^\top\ \theta_{\mathrm{dyn}}^\top\ \theta_{\mathrm{int}}^\top]^\top$ represents a vector of learned parameters that includes kinematics parameters, ten dynamic parameters of each pseudo-rigid body \cite{featherstone2014RBAlgos}, and interaction force parameters. 
In classical PRBA, $\tau_\mathrm{int}$ is modeled as a linear spring and damper torque:  $\tau_{\mathrm{int},i} = k_i q_{\prb,i} + c_i \dot q_{\prb,i}$, where $k_i, c_i \in \R^{2\times2}$ are diagonal matrices of stiffness and damping parameters, respectively \cite{ wittbrodt2007RFEM, moberg2014EFJ}. We propose a more versatile approach, illustrated in \figref{fig:DlOs}, by modeling the interaction forces using a general function approximator resulting in $\tau_{\mathrm{int}, i} = NN(q_{\prb,i}, \dot q_{\prb,i}, \theta_{\mathrm{int}, i})$ for $i$-th elastic joint.  

While being beyond the scope of this work,  $f_\mathrm{HybDyn}$ allows to incorporate external torques $\tau_\mathrm{ext}$ which arise from interactions with the environment. The torques can be readily computed as $\tau_\mathrm{ext} = \sum_i J_i(q_\prb, {}^\mathrm{b} p_i)^T F_i$ where $F_i$ denotes an external wrench, ${}^\mathrm{b} p_i$ represents the position at which the wrench is applied, and $J_i$ is its Jacobian. 
% Learning $\tau_\mathrm{ext}$ is challenging due to different combinations of ${}^\mathrm{b} p_i$ and $F_i$  potentially yielding identical torques. Yet, if $p_i$ is vaguely known using \eg tactile sensing \cite{sun2019machine}, then loss regularization may allow to uniquely determine ${}^\mathrm{b} p_i$ and $F_i$.

For simulation, we rewrite the second-order dynamics \eqref{eq:nprba_dynamics} in a standard state-space form: 
\begin{align} \label{eq:dyn_ode}
 \dot x = f(x,u, \theta) = [\dot q_{\prb}, f_\mathrm{HybDyn}(q_\prb, \dot q_\prb, \tau_\mathrm{int}, u, \theta)]  .
\end{align} 
Additionally, we define $x_{k+1} = F(x_k, u_{k}, \theta)$ as the one-step ahead prediction of states obtained through numerical integration of \eqref{eq:dyn_ode}.  For longer-term predictions, we consider $N$-step ahead predictions (with $N>1$) by recursively applying $F(\cdot)$ $N$ times which we refer to as a rollout.

\subsection{Initial state estimation} \label{subsec: state_est}
% Predicting DLO's motion via recursively calling $F(x_k, u_{k}, \theta)$
Calculating a rollout $x_{1}, x_{2} \dots, x_{N+1}$ requires an estimate of the initial state $x_0$. The simplest method to compute this initial state from inputs and outputs involves inverting the decoder (solving the inverse kinematics problem): $x_0 = h^{-1}(y_0, u_0)$. However, the kinematics of the PRBA is redundant, leading to non-uniqueness in the inverse kinematics solution. Therefore, we adopt a different approach during training in which we treat the initial states of all rollouts, denoted as $X_0$, as optimization variables. While this approach increases the number of decision variables, it effectively mitigates the non-uniqueness problem and enables the determination of initial states that are consistent with the dynamics.

When evaluating the model on test data, we compute each rollout's initial state using a method akin to moving horizon estimation \cite{rawlings2017MPCBook}. Concretely, to estimate state $x_0$ of each rollout, we use $m$ input and output samples prior to time instance $0$ to solve the following optimization problem 
\begin{align} \label{eq: state_est}
    \underset{x_{-m},\dots,x_{0}}{\min}\ &\frac{1}{m} \sum^{0}_{i=-m}  w_{\mathrm{est}, i} \|y_i - \hat y_i\|_2^2, \\
    \mathrm{s.t.}\ & x_{i+1} = F(x_{i}, u_{i}, \theta), \qquad i = -m,\dots, -1 \nonumber \\
    & \hat y_i = h(x_i, u_i), \ \qquad\qquad i = -m,\dots, 0 \nonumber
\end{align}
where $\hat y_i$ are output predictions of a trained model with fixed parameters starting from state $x_{-m}$. The weighting coefficients $w_{\mathrm{est}, i}$ enables the prioritization of certain samples.

\subsection{Model training} \label{subsec: model_train}
To jointly optimize the parameters $\theta$ of the model as well as the initial states $X_0$ of each rollout, we minimize a loss function consisting of several components, including MSE loss on output prediction and two regularization terms:
\begin{align}
    \underset{X_0, \theta}{\min}\ \frac{1}{n_r N} \sum_{i=1}^{n_r} \sum_{j=1}^N \|y_{i,j} - \hat y_{i,j} \|_{W_y}^2 + %\mathcal{L}_x +
    \mathcal{L}_{\mathrm{kin}} + \mathcal{L}_{\mathrm{dyn}}
\end{align}
where $n_r$ is the total number of rollouts, $N$ is the rollout length (the prediction horizon) and $W_y \in \R^{6\times6}$ is the diagonal weighting matrix. The regularization terms are defined as follows:
\begin{itemize}
    % \item $\mathcal{L}_x = \lambda_x \|X_0 - X_{0, \mathrm{IK}}\|_2^2$ regularizes the initial states with the solution on inverse kinematics $X_{0, \mathrm{IK}}$;
    \item $\mathcal{L}_{\mathrm{kin}} = \lambda_{\mathrm{kin}} \| \theta_{\mathrm{kin}} - \bar \theta_{\mathrm{kin}} \|_2^2$ regularizes kinematics parameters with uniform or heuristics-based discretization;
    \item $\mathcal{L}_{\mathrm{dyn}} = \lambda_q \|q_\prb \|_2^2 + \lambda_{\dot q} \|\dot q_\prb \|_2^2 + \lambda_{\mathrm{int}} \|\theta_\mathrm{int} \|_1$ regularizes the dynamics through kinetic and potential energy terms, and induces sparsity in the neural network weights. 
\end{itemize}

%===============================================================================
\section{Results} \label{sec:results}
To evaluate the proposed method, we predict the motion of two DLOs with significantly different properties as shown in the \figref{fig:DlOs}. The two DLO's are: i) an aluminum rod of 1.92\,m length, 4\,mm inner diameter and 6\,mm outer diameter, ii) a hollow polyethylene foam cylinder of 1.90\,m length and 60\,mm outer diameter. 
Our key finding is that NPRBA accurately predicts the DLOs dynamics while resorting to a low-dimensional state representation, and it allows for an approximate reconstruction of the entire shape of the DLOs. Moreover, it outperforms both black-box data-driven models and analytical PRBA model in the small data regime. The project code and data are available at: \url{https://tinyurl.com/neuralprba}.

\begin{figure}[h]
\begin{minipage}{.55\textwidth}
    \captionsetup{font=small}

    \scriptsize
    \setlength{\tabcolsep}{1.8pt}
    \begin{tabular}{ccc|cc}
         \toprule
         & \multicolumn{2}{c}{$\frac{1}{N}\sum||p_\mathrm{e} - \hat p_\mathrm{e}||_2$ [cm]} & \multicolumn{2}{c}{$\frac{1}{N}\sum||\dot p_\mathrm{e} - \hat{\dot p}_\mathrm{e}||_2$ [cm/s]} \\
         % & NPRBA (ours) & Vanilla-PRBA & PRBA & LTI & NODE  \\
         \cmidrule{2-5}
         & Foam cyl.\ & Alum.\ rod & Foam cyl.\ & Alum.\ rod \\
         \cmidrule{2-5} \\ 
         LTI & 18.4 $\pm$ 20.9 & 19.7$\pm$22.1 & 64.7$\pm$ 66.4& 120.3$\pm$133.2 \\
         NODE & 16.5$\pm$ 21.1 &  23.6$\pm$27.9 & 53.2 $\pm$ 58.2 &  152.6$\pm$167.1 \\
         {PRB-Net} \cite{mamedov2023FEIN}  & 15.0 $\pm$ 15.0 & 24.5$\pm$25.1 & 70.7$\pm$ 67.1 & 171.2 $\pm$ 181.2 \\
         % NODE-NN & ? & ? & ? & ? \\
         VPRBA & 5.8$\pm$ 9.3 & 12.7$\pm$ 8.2 & 21.2 $\pm$ 23.4 & 69.4 $\pm$ 41.1 \\
         % LPRBA & $5.0\pm 6.4$ & $12.0\pm 8.2$ & $18.7\pm 20.5$ & $64.8 \pm 34.8$ \\
         NPRBA (ours) & $\bm{4.3\pm 4.9}$ & $\bm{6.2\pm 4.6}$ & $\bm{15.3 \pm 15.0}$ & $\bm{36.5\pm28.3}$  \\
         % \midrule
         % FEIN \cite{mamedov2023FEIN} $\star$ & $3.7 \pm 3.1$ & $3.5\pm4.8$ & $17.4\pm14.7$ & $24.5 \pm 36.5$ \\
         % NRBA (ours) $\star$ & & $5.6\pm3.9$ & & $31.9 \pm 22.3$ \\
         \bottomrule
    \end{tabular}
    \captionof{table}{The one-second-ahead prediction accuracy on the test data for the the different DLO dynamics models.}
    \label{tab:accuracy_comparison}
\end{minipage}
\hfill
\begin{minipage}{.44\textwidth}
\captionsetup{font=small}
    \begin{center}
        \captionsetup{font=small}
        \includegraphics[width=\linewidth, trim={3mm 2mm 2mm 2mm}, clip]{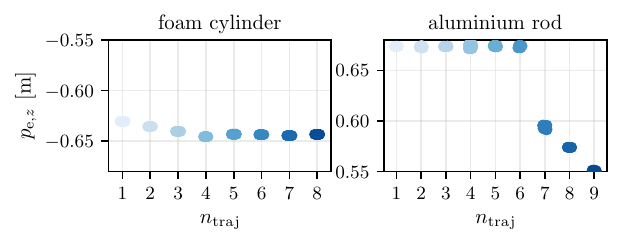}
        \caption{The DLO end positions along the $Z-$axis \emph{at rest} for different trajectories with the order of recording $n_{\text{traj}}$. 
        %A similar scale is chosen to highlight the changes in  $p_{\mathrm{e},z}$ between DLOs. 
        During the last trajectories, the metal rod got permanently deformed.}
        \label{fig:pe_z_at_rest}
    \end{center}
\end{minipage}
%\vspace{-6mm}
\end{figure}

\vspace{-2mm}
\subsection{Dataset} \label{subsec: dataset}

For learning the dynamics, we collected nine different trajectories for the aluminum rod and eight trajectories for the foam cylinder each containing about 30 seconds of data. Each trajectory started from a rest position, then during the first half (about 15 seconds) the robot arm actuated the DLO while during the second half the robot arm rested while the DLO kept moving. 
%The interested reader is referred to Appendix \ref{ap: data_collection} for more information on data collection and preprocessing.
\begin{comment}
\begin{wrapfigure}{r}{0.5\textwidth}
    \vspace{-5mm}
  \begin{center}
    \captionsetup{font=small}
    \includegraphics[width=0.48\textwidth, trim={3mm 0.2cm 0.1cm 0}]{images/train_trajs.pdf}
    \caption{Training trajectories of the DLO's end for the aluminum rod (Left) and foam cylinder (Right).}
    \label{fig:train_trajs}
  \end{center}
  \vspace{-4mm}
\end{wrapfigure}
\end{comment}

For training, only a single trajectory was used per DLO which accounts for 30 rollouts of 1 second  in the case of the aluminum rod and 23 rollouts of 1 second in the case of the foam cylinder. For testing, four trajectories with different frequency content and amplitude were used to analyze the generalizability of the models. More information on the dataset are provided in Appendix \ref{ap: data_collection}.

\paragraph{Plastic deformation} One notable property of the aluminum rod data is the distribution shift, which can be observed by comparing initial and final rest positions $p_\mathrm{e}$ for all the trajectories as depicted in \figref{fig:pe_z_at_rest}. 
%As shown in \figref{fig:pe_z_at_rest}, for the aluminum rod, $p_\mathrm{e}$ undergoes a noticeable change in $Z-$direction.
Given that the pose of frame $\{\mathrm{b}\}$ (see \figref{fig:notation} for reference) remained constant across experiments, the aluminum rod must have underwent permanent deformation, whereas the foam cylinder did not. More specifically, the initial deformation of the aluminum rod occurred during experiment six, due to the high accelerations. In subsequent experiments, which exhibited similar velocities and accelerations, additional deformations occurred. \Figref{fig:data_pn} and \figref{fig:data_ar} in Appendix \ref{ap: data_collection} show the train and test trajectories for both DLOs on a larger scale in terms of the DLO's end position $p_\mathrm{e}$, and the position $p_\mathrm{b}$ and orientation $\Psi_\mathrm{b}$ of the controlled end of the DLO. For the aluminum rod, the test trajectories include both trajectories before and after permanent deformations.
%Overall, the test trajectories contain different amplitudes to evaluate generalizability of the learned models.
%\blue{References to Appendix and Figures}

\subsection{Baselines}
We compared the proposed model with several common continuous-time models and with \cite{mamedov2023FEIN}, which also addresses DLO dynamics modeling:
\begin{enumerate}
    \item A linear time-invariant dynamics model (LTI);
    \item A second-order neural ordinary differential equation (NODE) \cite{chen2018NODE};
    \item {PRB-Net} \cite{mamedov2023FEIN} that uses a discrete-time recurrent neural network dynamics model%: gated recurrent unit (GRU);
    \item A vanilla PRBA model with a fixed heuristic discretization and optimized linear interaction torques and dynamical parameters (VPRBA).
\end{enumerate}
%
% \begin{figure}[t]
%     \vspace{-1mm}
%     \captionsetup{font=small}
%     \centering
%     \includegraphics[width=0.48\linewidth, trim={3mm 2mm 2mm 2mm}, clip]{images/acc_vs_nseg.pdf}
%     % \vspace{-6mm}
%     \caption{Influence of discretization (\eg number of pseudo-rigid bodies) on the prediction accuracy of NPRBA expressed as RMSE between the measured and predicted DLO's end position and velocity.}
%     \label{fig:discr_vs_acc}
%     \vspace{-3mm}
% \end{figure}
%
All models utilize the PRBA representation (where states are the generalized positions and velocities of the pseudo-joints) and the forward kinematics of the PRBA \eqref{eq: fk_decoder} as a decoder. More details on the model baselines are provided in Appendix \ref{apendix:baselines} while further training details are given in Appendix \ref{apendix:impl_training}.

\begin{comment}
\begin{wraptable}{r}{0.5\textwidth}
    \vspace{-5mm}
    \captionsetup{font=small}
    \scriptsize
    \setlength{\tabcolsep}{2.0pt}
  \begin{tabular}{cccccc}
        \toprule
         & Dyn. eq. & Dyn. type & Discr. & $\tau_\mathrm{int}$ \\
         \midrule\\
        LTI & $\dot x = Ax + Bu + c$ & black-box & optimized  & n/a\\
        NODE & $\dot x = NN(x, u)$& black-box & optimized & n/a\\
        PRB-Net \cite{mamedov2023FEIN} & $x_{k+1} = GRU(x, u)$ & black-box & optimized & n/a\\
        VPRBA & eq. \eqref{eq:dyn_ode}& grey-box & fixed & linear \\
        \bottomrule
    \end{tabular}
    \caption{Comparison of baseline models. All the models utilize forward kinematics of a PRBA as a decoder. }
    \label{tab:baselines}
  \vspace{-4mm}
\end{wraptable}
\end{comment}

\subsection{Experiments}
In what follows, we analyze how the NPRBA compares to the model baselines as well as the optimal number of pseudo-rigid bodies, inference time, and the model's ability to generalize to long horizon predictions. A detailed ablation study for the NPRBA is given in Appendix \ref{appendix:ablations}.

\subsubsection{Discretization selection}
\Figref{fig:results} (Right) illustrates the relationship between the number of PRBs $n_\prb$  and the prediction accuracy for the proposed model. The number of elastic joints equals $n_\mathrm{ej} = n_\prb - 1$ and defines the number of degrees of freedom which amount to $q_\prb \in \R^{2 n_\mathrm{ej}}$. After training the NPRBA for each DLO seperately, the proposed model reaches the highest accuracy with relatively few PRBs. For the foam cylinder, the position prediction accuracy drops below 5\,cm with just five PRBs. 
\begin{comment}
\begin{wrapfigure}{r}{0.45\textwidth}
    \vspace{-3mm}
   \begin{center}
    \captionsetup{font=small}
    \includegraphics[width=0.43\textwidth, trim={3mm 2mm 2mm 2mm}, clip]{images/acc_vs_nseg.pdf}
    \caption{Influence of discretization (\eg number of pseudo-rigid bodies) on the prediction accuracy of NPRBA expressed as RMSE between the measured and predicted DLO's end position and velocity.}
    \label{fig:discr_vs_acc}
  \end{center}
  \vspace{-4mm}
\end{wrapfigure}
\end{comment}
For the aluminum rod, even with three PRBs the position prediction accuracy reaches 6.1\,cm, just 0.1\,cm above the maximal accuracy (among all considered discretizations). The trend for velocity prediction accuracy mirrors that of position accuracy. Overall, among the two DLOs, the model more accurately predicts the motion of the foam cylinder which we attribute to the plastic deformation of the aluminium rod as detailed in Section \ref{subsec: dataset} %indicating that a better fit to the training data does not necessarily translate to a better fit in the test data. Another reason is 
and the varying amplitudes of motion: the standard deviation of $p_\mathrm{e}$ for the foam cylinder is $[0.09\ 0.35\ 0.09]$ cm, compared to $[0.13\ 0.44\ 0.25]$ cm for the aluminum rod. Furthermore, the foam cylinder is more highly damped rendering its ODE less stiff.

% In both cases, further refinement in discretization above five PRBs lead to zero or marginal improvement in prediction accuracy, but comes at a cost.
% %
% Models with a larger number of PRBs take longer to train, possess a higher-dimensional state space, and are more susceptible to overfitting compared to coarser models. Therefore, as a compromise between accuracy and state-space dimensionality, we selected models with five and three PRBs for the foam cylinder and aluminum rod, respectively, which we used in the subsequent experiments.

Adding more than five PRBs offers minimal to no improvement in prediction accuracy and introduces drawbacks such as longer training times, a larger state space, and increased risk of overfitting. For subsequent experiments, we chose models with five PRBs for the foam cylinder and three for the aluminum rod to balance accuracy with computational efficiency.

\subsubsection{Prediction accuracy}

\begin{figure}[t]
    \begin{minipage}{.7\textwidth}
    \centering
    \includegraphics[width=1.\textwidth]{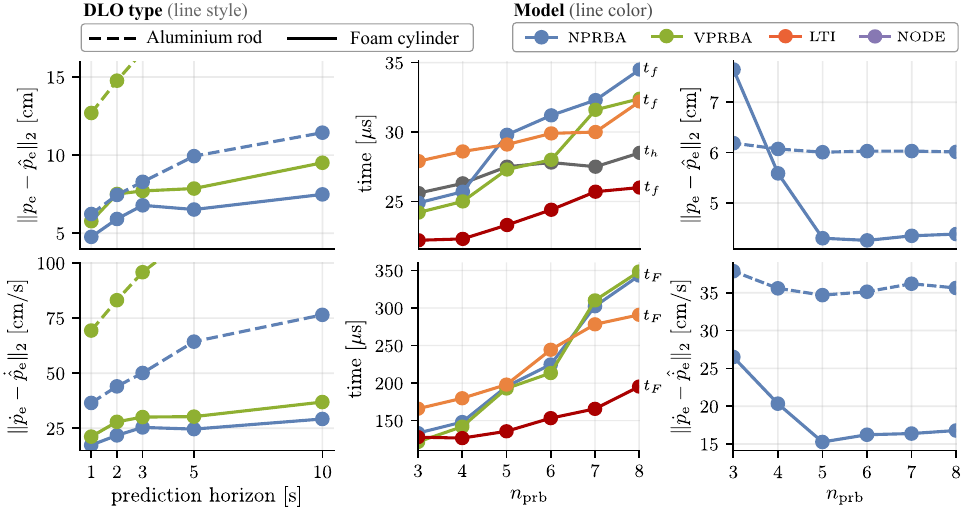}
    \caption{
    {\color{ourviolet2}\textbf{Left:}} Prediction error over varying prediction horizons. %fig:long_pred
    {\color{ourviolet2}\textbf{Center:}} Computation times of the continuous-time models for decoder evaluation $t_h$ (common for all models), continuous-time dynamics evaluation $t_f$, and the one-step integration $t_F$. %fig:t_inference
    {\color{ourviolet2}\textbf{Right:}} Influence of discretization (number of pseudo-rigid bodies) on NPRBA's RMSE between measured and predicted DLO's end state. %fig:discr_vs_acc
    }
    \label{fig:results}
    \end{minipage}
    \hfill
    \begin{minipage}{.28\textwidth}
    %\vspace{1cm}
    \captionsetup{font=small}
    \centering
    \includegraphics[width=\linewidth, trim={13.7cm 2mm 2mm 2mm}, clip]{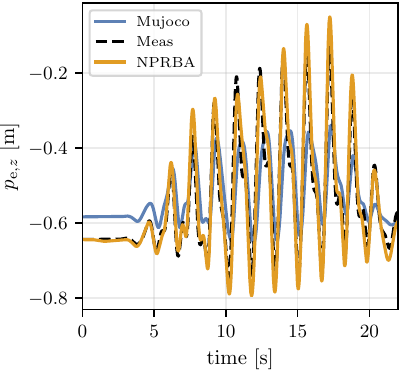}
    \caption{Comparison of NPRBA to a tuned Mujoco baseline model predicting the 8th foam cylinder trajectory starting from resting position. See also \figref{fig:ar_nprba_vs_mujoco} and \figref{fig:pn_nprba_vs_mujoco} for additional results.}
    \label{fig:prediction_example_small}
    \end{minipage}
\end{figure}

Table \ref{tab:accuracy_comparison} illustrates the prediction accuracy of both the proposed and baseline models  on test data, with a four segment discretization ($n_x = 16$) in the case of the foam cylinder, and two segment discretization ($n_x = 8$) in the case of the aluminum rod.
Notably, physics-based models demonstrate significantly superior performance compared to fully data-driven models like LTI, NODE and {PRB-Net} if these models are trained on small data sets.

Interestingly, for the foam cylinder the VPRBA with fixed heuristic-based discretization exhibits good performance. By optimizing discretization and incorporating learning of the nonlinear interaction forces, the NPRBA shows further improved prediction accuracy.
For the aluminum rod, VPRBA performed considerably worse than the proposed model as linear spring-damping forces are not sufficient to capture the dynamics if the DLO undergoes permanent deformations.  
%For a more detailed discussion, see Appendix \ref{apppendix:nprba_offset}. 
%In the result section of this work, we attributed the distribution shift in the data to permanent deformation of the DLO. 

\paragraph{Long prediction horizons}
To assess the generalizability of the top-performing models (NPRBA and VPRBA), we conducted rollouts with duration exceeding those used during training. While there is a decrease in prediction accuracy as the rollout length increases, NPRBA continues to outperform VPRBA, as shown in the \Figref{fig:results} (Left). \figref{fig:prediction_example_small} compares NPRBA to a finely-tuned Mujoco baseline simulation (see Appendix \ref{appendix:mujoco_comparison} for more details) predicting the full 8th foam cylinder trajectory starting from resting position. Notably, NPRBA correctly estimates the DLO end's initial state as well as the dynamic motion for a twenty second prediction horizon.
More comparisons of long horizon predictions are depicted in \figref{fig:ar_nprba_vs_mujoco} and \figref{fig:pn_nprba_vs_mujoco}.

% \begin{figure}
%     \captionsetup{font=small}
%     \centering
%     \includegraphics[width=\linewidth, trim={3mm 2mm 2mm 2.5mm}, clip]{images/inference_timings.pdf}
%     \caption{Computation times for the continuous-time models: $t_h$ is decoder/output evaluation time (common for all models), $t_f$ is continuous-time dynamics evaluation time and $t_F$ is one-step integration time. Evaluations were performed on a  laptop with an %Intel\textregistered~Core\texttrademark~i7-8550U CPU at 1.8~GHz processor.
%     Intel Core i7-8550U CPU at 1.8~GHz processor.}
%     \label{fig:t_inference}
%     \vspace{-4mm}
% \end{figure}

\subsubsection{Inference time}
Using an NPRBA model for motion planning and real-time control necessitates fast computation times for all its components, including the decoder, continuous-time dynamics, and one-step integration. To evaluate the model’s computation times, we adhered to the standard tool stack for model-based optimal control, implementing all models using CasADi \cite{Andersson2019casadi} and leveraging just-in-time compilation. For integration, we employed the casados integrator, designed specifically for real-time applications \cite{frey2023casados}. Evaluations were performed on an Intel Core i7-8550U CPU with 1.8~GHz processor.
\Figref{fig:results} (Center) displays the evaluation times of the decoder and dynamics evaluation, $t_h$ and $t_f$, respectively, which are in the order of microseconds and increase as spatial discretization becomes finer. To address the numerical stiffness of the underlying dynamics, a four-stage fixed-step implicit Runge-Kutta integrator was used. As shown in \figref{fig:results} (Center), this integration process remains sufficiently fast for real-time control applications, despite being an order of magnitude more costly than model evaluation.
Overall, among all continuous-time models, LTI is the fastest to evaluate and integrate. However, the proposed NPRBA maintains comparable computation times while offering better accuracy than VPRBA, NODE, and LTI.

% Using an NPRBA model for motion planning and real-time control requires fast computation times of all of its components, including the decoder, the continuous-time dynamics, and the one-step integration.
% To evalaute the model's computation times, we adhere to the common tool stack for model-based optimal control and implemented all models using CasADi \cite{Andersson2019casadi} while leveraging just-in-time compilation.
% For the integration, we use the casados integrator which has been designed for real-time applications  \cite{frey2023casados}. 

% Figure~\ref{fig:t_inference} shows the evaluation times of the decoder and dynamics evaluation, $t_h$ and $t_f$, respectively. The computation times are merely in the order of \emph{microseconds} and increase as spatial discretization becomes finer.
% To integrate \eqref{eq:dyn_ode}, a four-stage fixed-step implicit Runge-Kutta integrator was used due to the numerical stiffness of the underlying dynamics. Figure~\ref{fig:t_inference} shows that this integration remains sufficiently fast for real-time control applications even though integration is an order of magnitude more costly than model evaluation.
% Overall, among all the  continuous-time models LTI is the fastest to evaluate and integrate. The proposed NPRBA, while offering better accuracy than VPRBA, NODE and LTI, does not trade off the computation times.

%===============================================================================
\section{Conclusion} \label{sec:conclusion}
\begin{comment}
\begin{wrapfigure}{r}{0.45\textwidth}
    \vspace{-5mm}
  \begin{center}
    \captionsetup{font=small}
    \includegraphics[width=0.43\textwidth, trim={3mm 2mm 2mm 2mm}, clip]{images/generalization.pdf}
    \caption{Generalization of models measured in terms of position prediction accuracy on longer prediction horizons.}
    \label{fig:long_pred}
  \end{center}
  \vspace{-6mm}
\end{wrapfigure}
\end{comment}
This paper introduces a sample-efficient approach for modeling DLO dynamics, which integrates pseudo-rigid body approximation with neural networks. We replaced the traditional linear spring-damper with neural networks to model interaction torques and optimize the discretization of pseudo-rigid bodies. Our model, trained from a single trajectory, achieves high prediction accuracy on DLOs with varying physical properties, significantly outperforming both classical PRBA and black-box models, as well as Mujoco \cite{todorov2012mujoco} model with identified material properties from data.

%Our analysis reveals that optimizing discretization universally enhances model performance across different materials. For materials with relatively linear properties like aluminum, NPRBA's superior performance over classical PRBA is mainly due to its capacity to handle permanent deformations as sh. However, for materials exhibiting nonlinear behaviors, such as foam, the integration of neural networks in modeling interaction torques is essential for optimal results.

A limitation of NPRBA and other NeuralODEs are their longer training times compared to discrete-time models. However, with the development of faster and more robust numerical integrators, this aspect becomes less significant. Future research will explore the model's application in feedback control and extending its use to bi-manual manipulation tasks involving actuation of both DLO ends.
%Another avenue is to model plastic deformation as an offset in interaction torques, estimating this offset online to better address any occurring plastic deformations during manipulation.

%===============================================================================

\clearpage
% The acknowledgments are automatically included only in the final and preprint versions of the paper.
\acknowledgments{If a paper is accepted, the final camera-ready version will (and probably should) include acknowledgments. All acknowledgments go at the end of the paper, including thanks to reviewers who gave useful comments, to colleagues who contributed to the ideas, and to funding agencies and corporate sponsors that provided financial support.}

%===============================================================================

\bibliography{references}  % .bib

\begin{thebibliography}{33}
\providecommand{\natexlab}[1]{#1}
\providecommand{\url}[1]{\texttt{#1}}
\expandafter\ifx\csname urlstyle\endcsname\relax
  \providecommand{\doi}[1]{doi: #1}\else
  \providecommand{\doi}{doi: \begingroup \urlstyle{rm}\Url}\fi

\bibitem[Arriola-Rios et~al.(2020)Arriola-Rios, Guler, Ficuciello, Kragic, Siciliano, and Wyatt]{arriola2020DOmodelingSurvey}
V.~E. Arriola-Rios, P.~Guler, F.~Ficuciello, D.~Kragic, B.~Siciliano, and J.~L. Wyatt.
\newblock Modeling of deformable objects for robotic manipulation: A tutorial and review.
\newblock \emph{Frontiers in Robotics and AI}, 7:\penalty0 82, 2020.

\bibitem[Sanchez et~al.(2018)Sanchez, Corrales, Bouzgarrou, and Mezouar]{sanchez2018robotic}
J.~Sanchez, J.-A. Corrales, B.-C. Bouzgarrou, and Y.~Mezouar.
\newblock Robotic manipulation and sensing of deformable objects in domestic and industrial applications: a survey.
\newblock \emph{The International Journal of Robotics Research}, 37\penalty0 (7):\penalty0 688--716, 2018.

\bibitem[Yan et~al.(2020)Yan, Zhu, Jin, and Bohg]{yan2020biLSTMRope}
M.~Yan, Y.~Zhu, N.~Jin, and J.~Bohg.
\newblock Self-supervised learning of state estimation for manipulating deformable linear objects.
\newblock \emph{IEEE Robotics and Automation Letters}, 5\penalty0 (2):\penalty0 2372--2379, 2020.

\bibitem[Yang et~al.(2021)Yang, Stork, and Stoyanov]{yang2021biLSTM_IN}
Y.~Yang, J.~A. Stork, and T.~Stoyanov.
\newblock Learning to propagate interaction effects for modeling deformable linear objects dynamics.
\newblock In \emph{2021 IEEE International Conference on Robotics and Automation (ICRA)}, pages 1950--1957. IEEE, 2021.

\bibitem[Yu et~al.(2022)Yu, Lv, Zhong, Song, and Li]{yu2022RBF}
M.~Yu, K.~Lv, H.~Zhong, S.~Song, and X.~Li.
\newblock Global model learning for large deformation control of elastic deformable linear objects: An efficient and adaptive approach.
\newblock \emph{IEEE Transactions on Robotics}, 39\penalty0 (1):\penalty0 417--436, 2022.

\bibitem[Bauchau(2011)]{bauchau2011flexible}
O.~A. Bauchau.
\newblock \emph{Flexible multibody dynamics}, volume 176.
\newblock Springer, 2011.

\bibitem[Wittbrodt et~al.(2007)Wittbrodt, Adamiec-W{\'o}jcik, and Wojciech]{wittbrodt2007RFEM}
E.~Wittbrodt, I.~Adamiec-W{\'o}jcik, and S.~Wojciech.
\newblock \emph{Dynamics of flexible multibody systems: rigid finite element method}.
\newblock Springer Science \& Business Media, 2007.

\bibitem[Mamedov et~al.(2023)Mamedov, Geist, Swevers, and Trimpe]{mamedov2023FEIN}
S.~Mamedov, A.~R. Geist, J.~Swevers, and S.~Trimpe.
\newblock Pseudo-rigid body networks: learning interpretable deformable object dynamics from partial observations.
\newblock \emph{arXiv preprint arXiv:2307.07975}, 2023.

\bibitem[Lv et~al.(2017)Lv, Liu, Ding, Liu, Lin, and Ma]{lv2017physically}
N.~Lv, J.~Liu, X.~Ding, J.~Liu, H.~Lin, and J.~Ma.
\newblock Physically based real-time interactive assembly simulation of cable harness.
\newblock \emph{Journal of Manufacturing Systems}, 43:\penalty0 385--399, 2017.

\bibitem[Armanini et~al.(2023)Armanini, Boyer, Mathew, Duriez, and Renda]{armanini2023soft}
C.~Armanini, F.~Boyer, A.~T. Mathew, C.~Duriez, and F.~Renda.
\newblock Soft robots modeling: A structured overview.
\newblock \emph{IEEE Transactions on Robotics}, 2023.

\bibitem[Battaglia et~al.(2016)Battaglia, Pascanu, Lai, Jimenez~Rezende, et~al.]{battaglia2016interaction}
P.~Battaglia, R.~Pascanu, M.~Lai, D.~Jimenez~Rezende, et~al.
\newblock Interaction networks for learning about objects, relations and physics.
\newblock \emph{Advances in neural information processing systems}, 29, 2016.

\bibitem[Preiss et~al.(2022)Preiss, Millard, Yao, and Sukhatme]{preiss2022trackingPoolNoodle}
J.~A. Preiss, D.~Millard, T.~Yao, and G.~S. Sukhatme.
\newblock Tracking fast trajectories with a deformable object using a learned model.
\newblock In \emph{2022 International Conference on Robotics and Automation (ICRA)}, pages 1351--1357, 2022.

\bibitem[Li et~al.(2019)Li, Wu, Zhu, Tenenbaum, Torralba, and Tedrake]{li2019propNet}
Y.~Li, J.~Wu, J.-Y. Zhu, J.~B. Tenenbaum, A.~Torralba, and R.~Tedrake.
\newblock Propagation networks for model-based control under partial observation.
\newblock In \emph{ICRA}, 2019.

\bibitem[Moberg et~al.(2014)Moberg, Wernholt, Hanssen, and Brog{\aa}rdh]{moberg2014EFJ}
S.~Moberg, E.~Wernholt, S.~Hanssen, and T.~Brog{\aa}rdh.
\newblock Modeling and parameter estimation of robot manipulators using extended flexible joint models.
\newblock \emph{Journal of Dynamic Systems, Measurement, and Control}, 136\penalty0 (3):\penalty0 031005, 2014.

\bibitem[Bergou et~al.(2008)Bergou, Wardetzky, Robinson, Audoly, and Grinspun]{bergou2008discreterRods}
M.~Bergou, M.~Wardetzky, S.~Robinson, B.~Audoly, and E.~Grinspun.
\newblock Discrete elastic rods.
\newblock In \emph{ACM SIGGRAPH}, pages 1--12. 2008.

\bibitem[Buisson-Fenet et~al.(2022)Buisson-Fenet, Morgenthaler, Trimpe, and Di~Meglio]{buisson2022recognitionModels}
M.~Buisson-Fenet, V.~Morgenthaler, S.~Trimpe, and F.~Di~Meglio.
\newblock Recognition models to learn dynamics from partial observations with neural odes.
\newblock \emph{Transactions on Machine Learning Research}, 2022.

\bibitem[Chen et~al.(2018)Chen, Rubanova, Bettencourt, and Duvenaud]{chen2018NODE}
R.~T. Chen, Y.~Rubanova, J.~Bettencourt, and D.~K. Duvenaud.
\newblock Neural ordinary differential equations.
\newblock \emph{Advances in neural information processing systems}, 31, 2018.

\bibitem[Beintema et~al.(2023)Beintema, Schoukens, and T{\'o}th]{beintema2022DeepSubspace}
G.~I. Beintema, M.~Schoukens, and R.~T{\'o}th.
\newblock Continuous-time identification of dynamic state-space models by deep subspace encoding.
\newblock In \emph{The Eleventh International Conference on Learning Representations}, 2023.

\bibitem[Featherstone(2014)]{featherstone2014RBAlgos}
R.~Featherstone.
\newblock Rigid body dynamics algorithms.
\newblock 2014.

\bibitem[Rawlings et~al.(2017)Rawlings, Mayne, and Diehl]{rawlings2017MPCBook}
J.~B. Rawlings, D.~Q. Mayne, and M.~Diehl.
\newblock \emph{Model predictive control: theory, computation, and design}, volume~2.
\newblock Nob Hill Publishing Madison, WI, 2017.

\bibitem[Andersson et~al.(2019)Andersson, Gillis, Horn, Rawlings, and Diehl]{Andersson2019casadi}
J.~A.~E. Andersson, J.~Gillis, G.~Horn, J.~B. Rawlings, and M.~Diehl.
\newblock {CasADi} -- {A} software framework for nonlinear optimization and optimal control.
\newblock \emph{Mathematical Programming Computation}, 11\penalty0 (1):\penalty0 1--36, 2019.

\bibitem[Frey et~al.(2023)Frey, De~Schutter, and Diehl]{frey2023casados}
J.~Frey, J.~De~Schutter, and M.~Diehl.
\newblock Fast integrators with sensitivity propagation for use in casadi.
\newblock In \emph{2023 European Control Conference (ECC)}, pages 1--6. IEEE, 2023.

\bibitem[Todorov et~al.(2012)Todorov, Erez, and Tassa]{todorov2012mujoco}
E.~Todorov, T.~Erez, and Y.~Tassa.
\newblock Mujoco: A physics engine for model-based control.
\newblock In \emph{2012 IEEE/RSJ International Conference on Intelligent Robots and Systems}, pages 5026--5033. IEEE, 2012.

\bibitem[Bradbury et~al.(2018)Bradbury, Frostig, Hawkins, Johnson, Leary, Maclaurin, Necula, Paszke, Vander{P}las, Wanderman-{M}ilne, and Zhang]{jax2018github}
J.~Bradbury, R.~Frostig, P.~Hawkins, M.~J. Johnson, C.~Leary, D.~Maclaurin, G.~Necula, A.~Paszke, J.~Vander{P}las, S.~Wanderman-{M}ilne, and Q.~Zhang.
\newblock {JAX}: composable transformations of {P}ython+{N}um{P}y programs, 2018.
\newblock URL \url{http://github.com/google/jax}.

\bibitem[Kidger and Garcia(2021)]{kidger2021equinox}
P.~Kidger and C.~Garcia.
\newblock {E}quinox: neural networks in {JAX} via callable {P}y{T}rees and filtered transformations.
\newblock \emph{Differentiable Programming workshop at Neural Information Processing Systems}, 2021.

\bibitem[Kv{\ae}rn{\o}(2004)]{kvaerno2004kvaerno}
A.~Kv{\ae}rn{\o}.
\newblock Singly diagonally implicit runge--kutta methods with an explicit first stage.
\newblock \emph{BIT Numerical Mathematics}, 44\penalty0 (3):\penalty0 489--502, 2004.

\bibitem[Kidger(2021)]{kidger2021diffrax}
P.~Kidger.
\newblock \emph{{O}n {N}eural {D}ifferential {E}quations}.
\newblock PhD thesis, University of Oxford, 2021.

\bibitem[Loshchilov and Hutter(2019)]{loshchilov2017adamw}
I.~Loshchilov and F.~Hutter.
\newblock Decoupled weight decay regularization.
\newblock In \emph{International Conference on Learning Representations}, 2019.

\bibitem[Wensing et~al.(2017)Wensing, Kim, and Slotine]{wensing2017linear}
P.~M. Wensing, S.~Kim, and J.-J.~E. Slotine.
\newblock Linear matrix inequalities for physically consistent inertial parameter identification: A statistical perspective on the mass distribution.
\newblock \emph{IEEE Robotics and Automation Letters}, 3\penalty0 (1):\penalty0 60--67, 2017.

\bibitem[Sutanto et~al.(2020)Sutanto, Wang, Lin, Mukadam, Sukhatme, Rai, and Meier]{sutanto2020VirtualParams}
G.~Sutanto, A.~Wang, Y.~Lin, M.~Mukadam, G.~Sukhatme, A.~Rai, and F.~Meier.
\newblock Encoding physical constraints in differentiable newton-euler algorithm.
\newblock In \emph{Learning for Dynamics and Control}, pages 804--813. PMLR, 2020.

\bibitem[Elfwing et~al.(2018)Elfwing, Uchibe, and Doya]{elfwing2018siLU}
S.~Elfwing, E.~Uchibe, and K.~Doya.
\newblock Sigmoid-weighted linear units for neural network function approximation in reinforcement learning.
\newblock \emph{Neural networks}, 107:\penalty0 3--11, 2018.

\bibitem[Agrawal et~al.(2019)Agrawal, Amos, Barratt, Boyd, Diamond, and Kolter]{agrawal2019differentiable}
A.~Agrawal, B.~Amos, S.~Barratt, S.~Boyd, S.~Diamond, and J.~Z. Kolter.
\newblock Differentiable convex optimization layers.
\newblock \emph{Advances in neural information processing systems}, 32, 2019.

\bibitem[Book(1984)]{book1984AMM}
W.~J. Book.
\newblock Recursive lagrangian dynamics of flexible manipulator arms.
\newblock \emph{The International Journal of Robotics Research}, 3\penalty0 (3):\penalty0 87--101, 1984.

\end{thebibliography}

\newpage
\appendix
% \section{Appendix}
\section{Dataset: collection, preprocessing and analysis} \label{ap: data_collection}
To measure the motion of the DLO, two marker frames were attached to its start and end. These frames' motions were recorded by a `Vicon' camera-based positioning system. 
The raw sensor readings were measured in different frames (Vicon and Pandas base frames). To express measurements in the same frame, we calibrated the following constant transformations: the transformation from the Panda base frame $\{\mathrm{B}\}$ to the Vicon frame $\{\mathrm{V}\}$; the transformation from the Panda's end-effector frame to the DLO start frame $\{\mathrm{b}\}$; and the distance between the origins of frames $\{\mathrm{b}\}$ and $\{\mathrm{e}\}$ in the DLO's resting position. 

To excite the dynamics of the DLO, we employed the multisine trajectory as a reference for the Panda's joint velocity controller:
\begin{align}
    \dot{q}_\mathrm{p}(t) = \sum_{i=1}^{n_h} \alpha_i \omega_i \cos(\omega_i t) - \sum_{i=1}^{n_h} \beta_i \omega_i \sin(\omega_i t)
\end{align}
where $\alpha_i, \beta_i \in \mathbb{R}^{7}$ are the multisine coefficients, $n_h=6$ represents the number of harmonics, and $\omega_i$ are the excitation frequencies (harmonics). The coefficients $\alpha_i$ and $\beta_i$ were optimized through a constrained optimization problem, aiming to maximize the end-effector acceleration while remaining within the physical limits of the Panda robot. The harmonics in $\omega_i$ were selected experimentally based on a rough estimate of the DLO's first natural frequency. 

During data collection, the Panda's joint positions $ q_\mathrm{p}$ and velocities $\dot{ q}_\mathrm{p}$ as well as marker frame poses were recorded. During the data processing stage, the outputs $ y$ and inputs $ u$ were computed using the calibrated kinematics and forward kinematics of the robot arm.
As the Vicon system and the robot operate at different maximum sampling frequencies (Vicon system: $\sim 300$\,Hz, Robot arm: $\sim 600$\,Hz), to ensure that the data is sampled at the same frequency and to eliminate high-frequency noise, we processed the raw data using the following steps:
\begin{enumerate}
    \item Expressed the Vicon measurements in the robot base frame $\{\mathrm{B}\}$.
    \item Constructed a uniform grid with a sampling time of $\Delta t = 4$ ms.
    \item Designed a zero-phase low-pass Butterworth filter of order 8, sampling frequency $1/\Delta t$ and cut-off frequency $f_{\mathrm{co}} = 3.5 $ Hz. The filter's cut-off frequency was chosen empirically by observing data.
    \item For Panda robot data: (i) resampled $ q_\text{p}$ and $\dot{ q}_\mathrm{p}$ to a uniform grid and applied the designed filter; (ii) computed $\ddot{ q}_\mathrm{p}$ using central differences. For Vicon data: (i) resampled $ p_\mathrm{e}$ to a uniform grid and applied the designed filter; (ii) computed $\dot{ p}_\mathrm{e}$ using central differences.
    \item Computed pose $( p_\mathrm{b},\  \Psi_\mathrm{b})$ and its two time derivatives using the robot arm's forward kinematics.
\end{enumerate}

% \begin{figure}
%     \captionsetup{font=small}
%     \centering
%     \includegraphics[width=\linewidth, trim={3mm 2mm 2mm 2mm}, clip]{images/data_dist_at_rest.pdf}
%     \caption{The DLOs end positions along the $Z-$axis \emph{at rest} for different recorded trajectories, with the order of recording preserved. A similar scale is chosen to highlight the changes in  $p_{\mathrm{e},z}$ between DLOs. For the last trajectories of the aluminum rod, strong excitation caused permanent deformation.}
%     \label{fig:pe_z_at_rest}
%     % \vspace{-4mm}
% \end{figure}

\begin{figure}[t]
    \captionsetup{font=small}
    \centering
    \includegraphics[width=\textwidth, trim={3mm 2mm 2mm 2mm}, clip]{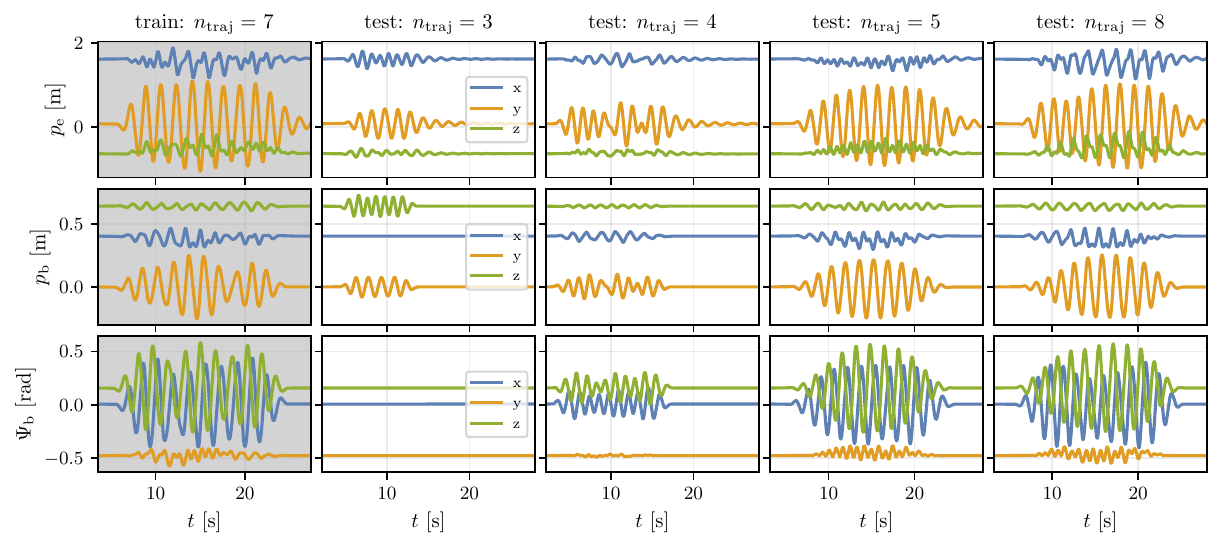}
    \caption{Foam cylinder train and test trajectories. For each trajectory, the number $n_{\mathrm{traj}}$, indicating in which sequence data has been collected (see \figref{fig:pe_z_at_rest}), is shown. }
    \label{fig:data_pn}
    \vspace{-4mm}
\end{figure}

\begin{figure}[t]
    \captionsetup{font=small}
    \centering
    \includegraphics[width=\textwidth, trim={3mm 2mm 2mm 2mm}, clip]{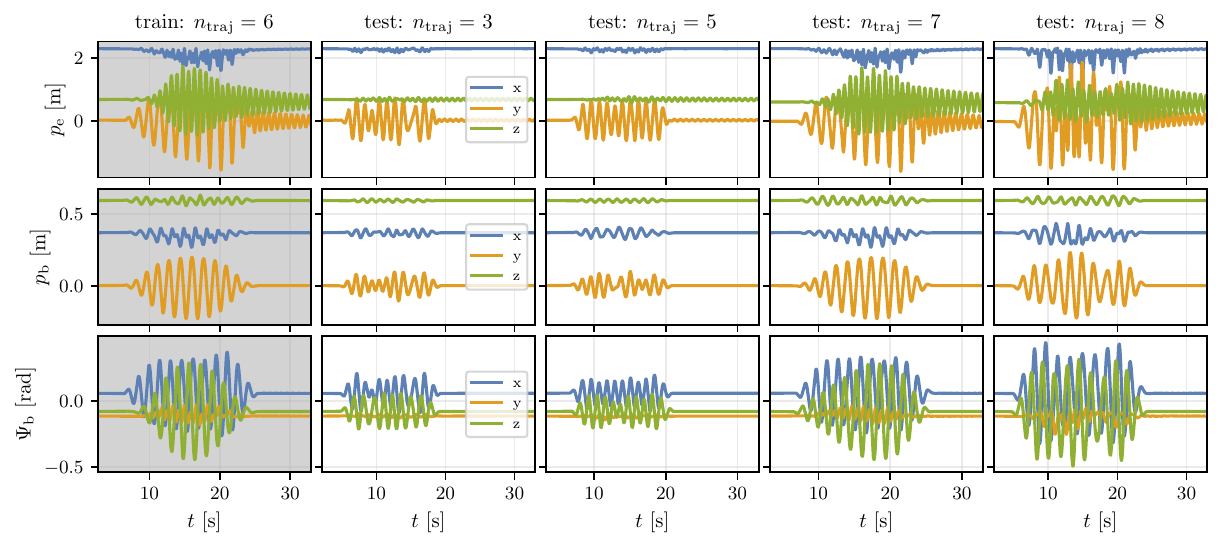}
    \caption{Aluminium rod train and test trajectories. For each trajectory, the number $n_{\mathrm{traj}}$, indicating in which sequence data has been collected (see \figref{fig:pe_z_at_rest}), is shown.}
    \label{fig:data_ar}
    \vspace{-4mm}
\end{figure}

\section{Baseline Models}\label{apendix:baselines}
\begin{wraptable}{r}{0.5\textwidth}
    \vspace{-3mm}
    \captionsetup{font=small}
    \scriptsize
    \setlength{\tabcolsep}{2.0pt}
  \begin{tabular}{cccccc}
        \toprule
         & Dyn. eq. & Dyn. type & Discr. & $\tau_\mathrm{int}$ \\
         \midrule\\
        LTI & $\dot x = Ax + Bu + c$ & black-box & optimized  & n/a\\
        NODE & $\dot x = NN(x, u)$& black-box & optimized & n/a\\
        PRB-Net \cite{mamedov2023FEIN} & $x_{k+1} = GRU(x, u)$ & black-box & optimized & n/a\\
        VPRBA & eq. \eqref{eq:dyn_ode}& grey-box & fixed & linear \\
        \bottomrule
    \end{tabular}
    \caption{Comparison of baseline models. All the models utilize forward kinematics of a PRBA as a decoder. }
    \label{tab:baselines}
  \vspace{-4mm}
\end{wraptable}
The baselines were selected based on their frequent usage in different scientific disciplines that focus on dynamics identification: LTI denotes a common model in system identification; GRU  and NODE are standard models in machine learning; and VPRBA is frequently used in robotics. 
The LTI model acts as a sanity check that demonstrates that linear dynamics cannot adequately capture DLO motion.
From a physics point of view, LTI models can be interpreted as a linearization of \eqref{eq:dyn_ode} around a state-input pair $(x^\star, u^\star)$,
\begin{align}
    \dot x &\approx  \begin{bmatrix}
        0 & I \\
        \frac{\partial f_\mathrm{HybDyn}}{\partial q_\prb} & \frac{\partial f_\mathrm{HybDyn}}{\partial \dot q_\prb}
    \end{bmatrix} x + 
    \begin{bmatrix}
        0 \\ \frac{\partial f_\mathrm{HybDyn}}{\partial u}
    \end{bmatrix} u + f(x^\star, u^\star) \nonumber \\
    &= A(x^\star, u^\star) x + B(x^\star, u^\star) u + c(x^\star, u^\star),
\end{align}
where the matrices $A$ and $B$, and the vector $c$ are learned from data. 
NODE and {GRU} are both black-box NN models; however, NODE is a continuous-time model that treats initial states as optimization variables, while {GRU} is a discrete-time model and GRU-based PRB-Net estimates initial states using a recognition model. The distinctions among these baseline models are summarized in Table \ref{tab:baselines}.

\section{Implementation and Training of Model Baselines} \label{apendix:impl_training}
All models were implemented in JAX \cite{jax2018github}, using the Equinox library \cite{kidger2021equinox}. 
The numerical integration was performed using the implicit Runge-Kutta method \cite{kvaerno2004kvaerno} with a adaptive step-size from Diffrax library \cite{kidger2021diffrax}. For optimization, the batch gradient descent algorithm AdamW \cite{loshchilov2017adamw} was employed. To ensure physical consistency constraints for rigid body dynamic parameters \cite{wensing2017linear}, reparameterization was performed in accordance with \cite{sutanto2020VirtualParams}.
Nonlinear internal force was implemented as a linear model complemented with a residual NN with a single hidden layer of 8 nodes and SiLU activation function \cite{elfwing2018siLU}:  
\begin{align*}
    \tau_{\mathrm{int}, i} = &k_i q_{\prb,i} + c_i \dot q_{\prb,i} + W_{i,2} \cdot \left( \mathrm{SiLU}\left(W_{i,1}\cdot [q_{\prb,i}, \dot q_{\prb,i}]\right) + b_{i,2} \right).
\end{align*} 
In the cost function, the kinematics parameters of the aluminum  rod  were regularized with a uniform spatial discretization: $\bar \theta_\mathrm{kin} = [l_{\prb,1} \dots l_{\prb,n_\prb}]^\top$ where $ l_{\prb,i} = L/n_\prb$. For the foam cylinder uniform discretization is a poor choice because most of the deformation is distributed close to frame $\{\mathrm{b}\}$ (see \figref{fig:notation}). In this case, we regularize kinematics parameters using the heuristic: $\bar \theta_\mathrm{kin} = [0.1\ \dots \ 0.1\ L-0.1(n_\prb - 1)]^T$. 

\begin{wraptable}{r}{0.5\textwidth}
    \vspace{-4mm}
    \captionsetup{font=small}
    \centering
    \scriptsize
    \setlength{\tabcolsep}{3.5pt}
    \begin{tabular}{cccccc}
        \toprule
         & LTI & NODE & PRB-Net \cite{mamedov2023FEIN} & VPRBA & NPRBA \\
         \midrule
         foam cylinder, [s] & $1.0$ & $1.6$ & $2.9$ & $25.5$ & $26.7$ \\
         aluminum rod, [s] & $1.5$ & $4.4$ & $1.8$ & $21.4$ & $20.4$ \\
         \bottomrule
    \end{tabular}
    \caption{Training time for models over a single epoch, with 23 samples for the foam cylinder and 30 samples for the aluminum rod, was conducted on a laptop equipped with an NVIDIA GeForce MX250 GPU.}
    \label{tab:training_time}
  \vspace{-3mm}
\end{wraptable}

In case of NPRBA, simultaneously optimizing segment lengths, pseudo-rigid body parameters and interaction forces often converges to a bad local minima. We found that sequential approach yields better results. Concretely, we first fix the segment lengths, allowing the optimization of dynamic and interaction force parameters to reach a good local minimum. Thereafter, we perform a full parameter optimization.

The models require different amounts of time for training which depends on the complexity of the dynamics, the integrator, and the backpropogation algorithm.  Table~\ref{tab:training_time} shows that NPRBA and VPRBA take substantially more time to train than other baseline models.

% \begin{table}[b]
%     \captionsetup{font=small}
%     \centering
%     \scriptsize
%     \begin{tabular}{ccc|cc}
%         \toprule
%          & \multicolumn{2}{c}{$\frac{1}{N}\sum||p_\mathrm{e} - \hat p_\mathrm{e}||_2$ [cm]} & \multicolumn{2}{c}{$\frac{1}{N}\sum||\dot p_\mathrm{e} - \hat{\dot p}_\mathrm{e}||_2$ [cm/s]} \\
%          % & NPRBA (ours) & Vanilla-PRBA & PRBA & LTI & NODE  \\
%          \cmidrule{2-5}
%          & FC & AR & FC & AR \\
%          \cmidrule{2-5} \\ 
%        NPRBA &   $4.3\pm 4.9$ & $6.2\pm 4.6$ & $15.3 \pm 15.0$ & $36.5\pm28.3$ \\
%        Lin. $\tau_\mathrm{int}$ & $5.0 \pm 6.4$ &  $12.0\pm7.8$ & $18.7 \pm 20.5$ &  $63.1\pm 34.4$ \\
%        Aff. $\tau_\mathrm{int}$  & $4.7\pm 5.8$ & $6.3\pm 4.6$  & $17.0 \pm 18.1$ & $37.2 \pm 28.4$ \\
%        Discr. & $4.6\pm 4.8$ & $6.8\pm 4.9$ & $15.7 \pm 14.9$ & $39.6 \pm 34.3$ \\
%        % VPRBA & $5.8\pm 9.3$ & $12.7\pm 8.2$ & $21.2 \pm 23.4$ & $69.4 \pm 41.1$ \\
%        Data quality  & $7.3 \pm 12.4$  & $11.5 \pm 13.2$ & $26.4 \pm 38.0$ & $78.3 \pm 104.9$ \\
%        Data $\times 0.5$  & $5.4 \pm 6.4$  & $6.5 \pm 4.6$ & $18.9 \pm 19.5$ & $38.6 \pm 27.1$ \\
%        Data $\times 0.3$  & $6.4 \pm 5.7$  & $6.9 \pm 4.7$ & $19.7\pm 16.3$ & $40.4 \pm 29.5$ \\
%        $X_0$ & $5.1\pm 5.6$ & $7.0 \pm 5.0$ & $18.9 \pm 18.1$ & $42.0 \pm 35.9$ \\
%        % $2\times$ data  & $4.3 \pm 5.5$ & $6.1 \pm 5.4$ & $16.3 \pm 18.2$ & $35.8 \pm 33.5$\\
%        \bottomrule
%     \end{tabular}
%     \caption{Results of the NPRBA ablation studies. }
%     \label{tab:ablation}
%     \vspace{-4mm}
% \end{table}

\section{Ablation studies} \label{appendix:ablations}
% This subsection analyses the contribution of two main components of the proposed model -- nonlinear interaction torques and discretization optimization -- to its overall performance. In addition, it analyses the impact of the data quality, and a method for obtaining initial states on NPRBA performance. 
This subsection examines how two key components — nonlinear interaction torques and discretization optimization — along with data quality and initial state estimation, affect NPRBA's overall performance.
\subsection{Nonlinear interaction torques} To analyze the impact of the nonlinear interaction torques, which are a key component of the NPRBA, we trained a model where we replaced them with a linear $\tau_\mathrm{int}$. As shown in Table \ref{tab:ablation} (row \emph{Lin. $\tau_\mathrm{int}$}), the absence of nonlinear $\tau_\mathrm{int}$ significantly decreases NPRBA's performance, particularly for the aluminum rod.

\begin{wraptable}{r}{0.55\textwidth}
    % \vspace{-2mm}
    \captionsetup{font=small}
    \centering
    \scriptsize
    \setlength{\tabcolsep}{2.5pt}
    \begin{tabular}{ccc|cc}
        \toprule
         & \multicolumn{2}{c}{$\frac{1}{N}\sum||p_\mathrm{e} - \hat p_\mathrm{e}||_2$ [cm]} & \multicolumn{2}{c}{$\frac{1}{N}\sum||\dot p_\mathrm{e} - \hat{\dot p}_\mathrm{e}||_2$ [cm/s]} \\
         % & NPRBA (ours) & Vanilla-PRBA & PRBA & LTI & NODE  \\
         \cmidrule{2-5}
         & Foam cyl.\ & Alum.\ rod & Foam cyl.\ & Alum.\ rod \\
         \cmidrule{2-5} \\ 
       NPRBA &   $4.3\pm 4.9$ & $6.2\pm 4.6$ & $15.3 \pm 15.0$ & $36.5\pm28.3$ \\
       Lin. $\tau_\mathrm{int}$ & $5.0 \pm 6.4$ &  $12.0\pm7.8$ & $18.7 \pm 20.5$ &  $63.1\pm 34.4$ \\
       Aff. $\tau_\mathrm{int}$  & $4.7\pm 5.8$ & $6.3\pm 4.6$  & $17.0 \pm 18.1$ & $37.2 \pm 28.4$ \\
       Discr. & $4.6\pm 4.8$ & $6.8\pm 4.9$ & $15.7 \pm 14.9$ & $39.6 \pm 34.3$ \\
       % VPRBA & $5.8\pm 9.3$ & $12.7\pm 8.2$ & $21.2 \pm 23.4$ & $69.4 \pm 41.1$ \\
       Data quality  & $7.3 \pm 12.4$  & $11.5 \pm 13.2$ & $26.4 \pm 38.0$ & $78.3 \pm 104.9$ \\
       Data $\times 0.5$  & $5.4 \pm 6.4$  & $6.5 \pm 4.6$ & $18.9 \pm 19.5$ & $38.6 \pm 27.1$ \\
       Data $\times 0.3$  & $6.4 \pm 5.7$  & $6.9 \pm 4.7$ & $19.7\pm 16.3$ & $40.4 \pm 29.5$ \\
       $X_0$ & $5.1\pm 5.6$ & $7.0 \pm 5.0$ & $18.9 \pm 18.1$ & $42.0 \pm 35.9$ \\
       % $2\times$ data  & $4.3 \pm 5.5$ & $6.1 \pm 5.4$ & $16.3 \pm 18.2$ & $35.8 \pm 33.5$\\
       \bottomrule
    \end{tabular}
    \caption{Results of the NPRBA ablation studies. }
    \label{tab:ablation}
  \vspace{-3mm}
\end{wraptable}

However, an analysis of the weights and output of the NN modeling interaction torques reveals an interesting insight: the main difference from linear interaction torques is the presence of an offset. This observation is particularly easy to explain in the case of the aluminum rod: the offset captures the average material deformations. To test whether adding an offset to linear interaction torques, resulting in affine $\tau_\mathrm{int}$, could improve prediction accuracy, we trained a model with this modification. The results, presented in Table \ref{tab:ablation} (row \emph{Aff. $\tau_\mathrm{int}$}), show that adding an offset significantly improves performance, especially for the aluminum rod, nearly matching the accuracy of NPRBA.

This ablation study indicates that the aluminum rod typically exhibits linear behavior. However, in real-world scenarios, deformable objects may undergo permanent deformations. Such deformations can be effectively modeled by introducing a simple offset, leading to affine interaction torques. 
However, in the case of the foam cylinder, simply adding an offset does not achieve the same level of improved accuracy as for the aluminum rod, indicating that for this material the nonlinear component of the interaction torques plays a more important role. This highlights the contribution of the NN, as it is capable of capturing this nonlinear component in the interaction torques.
% Yet, the observation that merely adding an offset does not achieve results comparable to NPRBA for the foam cylinder highlights the importance of the NN. This is because the NN is capable of capturing a wide range of nonlinearities in the interaction torques.

\subsection{Discretization}
To quantify the impact of optimizing discretization (the lengths of pseudo-rigid bodies), we trained NPRBA with a fixed discretization, the same as is used for VPRBA.  
For both DLOs, using a fixed discretization led to a notable decrease in performance, as reported in Table \ref{tab:ablation} (row \emph{Discr.}).  
This result emphasizes the critical role of discretization optimization in the sample-efficient modeling of DLOs.
%Comparing this outcome with the previous ablation study indicates that optimizing the discretization contributes more to prediction accuracy than introducing nonlinearities.

\subsection{Data quality} 
In the small data regime, selecting the appropriate trajectory for training is crucial. The chosen trajectory must sufficiently excite the dynamics of the DLO and cover parts of the state-space relevant to downstream tasks, such as manipulation. For optimizing both baseline and NPRBA parameters, the training trajectory was carefully selected. To demonstrate the impact of trajectory quality, we trained NPRBA using a less exciting trajectory, one whose amplitude does not contain the full range present in the test data. As shown in Table \ref{tab:ablation} (row \emph{Data quality}), this led to a significance decline in performance when evaluated against the test data. This outcome underscores the significance of using high-quality training data that is sufficiently exciting and representative of the trajectories encountered in downstream tasks.

\subsection{Data quantity}
To evaluate how performance varies with dataset size, we trained NPRBA using just 50\,\% and 30\,\% of the training trajectory. As shown in Table~\ref{tab:ablation} (rows \emph{Data $\times 0.5$ and Data $\times 0.3$}), performance decreases with smaller datasets.  Notably, the foam cylinder model declines more than the aluminum rod, likely due to its material's higher nonlinearity. Despite using only half the data, NPRBA still outperforms VPRBA and other black-box models, which used the full trajectory for training, as detailed in Table~\ref{tab:accuracy_comparison}. In case of the aluminum rod, even with third of data NPRBA outperforms VPRBA and the black-box models.

\subsection{Initial state} 
Initial states $X_0$ play a significant role in accurately predicting the motion of DLOs over short horizons. Although it is possible to obtain $X_0$ by solving the inverse kinematics (IK) problem, this method has two limitations: the solution is not unique, and the formulation does not directly consider the flexibility of the joints. Nevertheless, IK can serve as a simple encoder and reduce the number of optimization variables. To test this hypothesis, we trained NPRBA using the solution from the IK problem. For simplicity, we fixed the discretization as in VPRBA, as implementing IK as an implicit layer and backpropagating through it would be complex \cite{agrawal2019differentiable}. The results, as shown in Table \ref{tab:ablation} (row $X_0$), indicate that IK is a strong candidate for an encoder. We hypothesize that if implemented as an implicit layer, it could achieve performance comparable to directly optimizing $X_0$.

\section{On other classical analytical methods} \label{appendix:mujoco_comparison}
% \begin{figure}
%     \captionsetup{font=small}
%     \centering
%     \includegraphics[width=0.87\linewidth]{images/classical_methods.pdf}
%     \caption{Simple planar example demonstrating the difference between small and large deformations for a DLO of length $L$. Euler-Bernoulli beam theory assumes small deformations -- $p_x^\mathrm{b}(t) = [x\ w(x,t)]^T$ and $p_\mathrm{e}^\mathrm{b}(t) = [L\ w(x,t)]^T$ -- which is suitable for fairly rigid robots but not for DLOs that exhibit large deformations, such as the foam cylinder.}
%     \label{fig:deformations}
%     \vspace{-4mm}
% \end{figure}

\begin{wrapfigure}{r}{0.45\textwidth}
    \vspace{-3mm}
    \begin{center}
        \captionsetup{font=small}
        \includegraphics[width=0.43\textwidth]{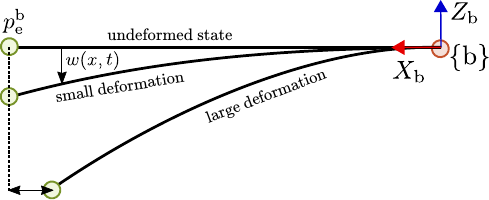}
        \caption{Simple planar example demonstrating the difference between small and large deformations for a DLO of length $L$. 
        %Euler-Bernoulli beam theory assumes small deformations which is suitable for fairly rigid robots but not for DLOs that exhibit large deformations, such as the foam cylinder.
        }
        \label{fig:deformations}
    \end{center}
  \vspace{-6mm}
\end{wrapfigure}

Euler-Bernoulli theory is a powerful method for modeling deformations of long slender beams. The theory makes an important assumption: deformations of a beam are small as demonstrated in \figref{fig:deformations}, thereby the deformation along $X_\mathrm{b}$ is ignored and position of a point along the DLO is described as $p_x^\mathrm{b} = [x\ w(x,t)]^T$. Although Euler-Beam theory has been applied for modeling the dynamics of flexible robots \cite{book1984AMM}, it is less suitable for DLOs undergoing large deformations. For this case, geometrically exact methods were developed that result in partial differential equations \cite{bauchau2011flexible}. The most common approaches for approximately solving these equations include finite element methods \cite{bauchau2011flexible} and discrete elastic rods \cite{bergou2008discreterRods}. The latter provides a good trade-off between accuracy and computation time. Mujoco \cite{todorov2012mujoco} offers an implementation of a variation of discrete elastic rods for modeling DLOs. We use this implementation as an additional analytical baseline by modeling both DLOs in Mujoco. 

\begin{figure*}[t]
    \captionsetup{font=small}
    \centering
    \includegraphics[width=\textwidth, trim={3mm 2mm 2mm 2mm}, clip]{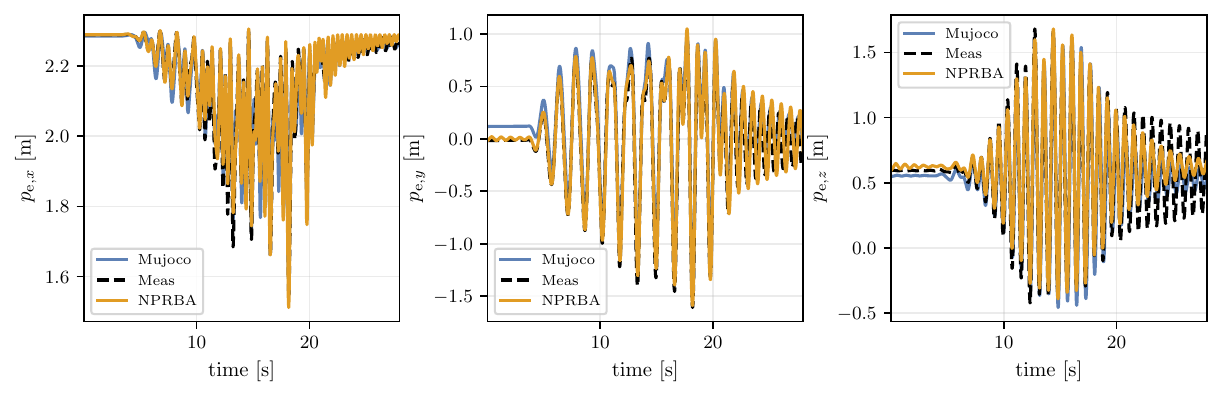}
    \caption{A comparison of the entire trajectory prediction ($n_\mathrm{traj}=7$) starting from a rest state for the aluminum rod using the proposed method and Mujoco.}
    \label{fig:ar_nprba_vs_mujoco}
\end{figure*}

\begin{figure*}[t]
    \captionsetup{font=small}
    \centering
    \includegraphics[width=\textwidth, trim={3mm 2mm 2mm 2mm}, clip]{images/pn_nprba_vs_mujoco.pdf}
    \caption{A comparison of the entire trajectory prediction ($n_\mathrm{traj}=8$) starting from a rest state for the foam cylinder using the proposed method and Mujoco.}
    \label{fig:pn_nprba_vs_mujoco}
\end{figure*}

For modeling DLOs, we specified their geometry, spatial discretization in terms of the number of segments $n_\mathrm{seg}$, and material properties: density $\rho$, Young's modulus $E$, and damping $c$. Initial values for these material properties were taken from available catalogs and subsequently fine-tuned. To accomplish this, we utilized the entire train trajectory and employed finite-difference methods for estimating gradients, as Mujoco does not currently provide gradient calculations and the parameter space is small. The final parameters used for the simulation are listed in  Table \ref{tab:mujoco_params}.
\begin{wraptable}{r}{0.5\textwidth}
    \vspace{-2mm}
    \captionsetup{font=small}
    \centering
    \scriptsize
    \setlength{\tabcolsep}{3.5pt}
    \begin{tabular}{ccccc}
        \toprule
         & $\rho$ [kg/m${}^3$] & $E$ [Pa] & $c$ [N$\cdot$ m $\cdot$ s/rad] & $n_\mathrm{seg}$ \\
         \midrule
       foam cylinder  & $1.93\times 10^2$ & $1.50\times 10^7$ & $1.93$ & $21$\\
       aluminum rod & $2.20\times 10^3$ & $5.95\times 10^{10}$ & $0.36$ & $21$ \\
         \bottomrule
    \end{tabular}
    \caption{ Spatial discretization and material properties of the DLOs models used for simulation in Mujoco. A large number of segments $n_\mathrm{seg}$ was chosen to enhance accuracy.}
    \label{tab:mujoco_params}
  \vspace{-3mm}
\end{wraptable}

The most straightforward way to compare the Mujoco DLO model with NPRBA is by predicting the full trajectory starting from a rest position, due to the gradient limitations mentioned previously and the large state dimension. When evaluated across all test trajectories, NPRBA consistently outperforms Mujoco, even though it was initially trained for one-step-ahead prediction and employs significantly coarser spatial discretization. Mujoco demonstrates similar prediction accuracies, measured by the RMSE of $p_\mathrm{e}$, for both DLOs: 18 cm for the aluminum rod and 19 cm for the foam cylinder. However, NPRBA significantly outperforms Mujoco, especially in predicting the motion of the foam cylinder with an RMSE of 7 cm, and also shows better performance for the aluminum rod with an RMSE of 15 cm. Examples of the predicted trajectories are shown in \figref{fig:ar_nprba_vs_mujoco} for the aluminum rod and in \figref{fig:pn_nprba_vs_mujoco} for the pool noodle.

\end{document}